\documentclass{article}

\PassOptionsToPackage{numbers}{natbib}
 \usepackage[preprint]{neurips_2025}

\usepackage[utf8]{inputenc} 
\usepackage[T1]{fontenc}    
\usepackage{hyperref}       
\usepackage{url}            
\usepackage{booktabs}       
\usepackage{amsfonts}       
\usepackage{nicefrac}       
\usepackage{microtype}      
\usepackage{xcolor}         
\usepackage{graphicx}
\usepackage{amsmath}
\usepackage{algorithm}
\usepackage{algorithmic}

\newcommand{\gap}{\mathcal{G}}
\newcommand{\fp}{\mathrm{FP32}}
\newcommand{\qt}{\mathrm{INT4}}
\newcommand{\ppl}{\mathrm{PPL}}

\title{When Flat Minima Fail: Characterizing INT4 Quantization
Collapse After FP32 Convergence}

\author{%
  Marcus Armstrong\\
  Department of Computer Science\\
  University of Houston\\
  Houston, TX  77004\\
  \texttt{miarmstr@cougarnet.uh.edu} \\
}

\begin{document}

\maketitle

\begin{abstract}
Post-training quantization (PTQ) assumes that a well-converged
model is a quantization-ready model. We show this assumption
fails in a structured, measurable, and previously uncharacterized
way. Using a calibration-free per-group INT4 probe applied to all
154 publicly available Pythia-160m training checkpoints, we
identify a three-phase divergence structure: a rapid-learning
phase where both FP32 perplexity and quantization robustness
improve together, a meta-stable plateau lasting roughly 70{,}000
steps where FP32 perplexity stagnates but INT4 gap remains
bounded, and an explosive divergence phase where the INT4 gap
compounds from 11\% to 517\% while FP32 perplexity barely
moves. Critically, this divergence begins not when the learning
rate starts decaying, but precisely when FP32 perplexity
converges---a finer-grained onset predictor that implies
post-convergence weight updates, rather than decay magnitude
alone, are the proximate cause. We further show that INT8
quantization is entirely immune throughout all three phases,
constraining the mechanism to the coarseness of the 16-level
INT4 grid specifically, and rule out weight outlier accumulation
as the mechanism via direct kurtosis measurement. Finally, we
conduct a controlled fork experiment from the pre-divergence
checkpoint comparing three learning rate schedules
(cosine continuation, SGDR warm restarts, and our proposed
Oscillatory Lock-In) across nine independent runs. SGDR
uniformly accelerates divergence (0/9 pairwise wins against
cosine), while OLI's settled cool phases reduce the INT4 gap
by 2.2 percentage points on average ($t = -5.46$,
$p < 0.0001$), demonstrating that schedule \emph{amplitude
calibration}, not oscillation alone, determines whether
perturbation helps or hurts. Our code, probe implementation,
and all 154-checkpoint audit results are released publicly.
\end{abstract}

\section{Introduction}
\label{sec:intro}
 
The transition from full-precision training to low-precision
inference has become one of the most practically consequential
operations in modern deep learning deployment. Post-training
quantization (PTQ) methods~\cite{frantar2023gptqaccurateposttrainingquantization,
xiao2024smoothquantaccurateefficientposttraining, lin2024awq} compress a trained model without
retraining, making them attractive for rapid deployment of large
language models (LLMs). Their correctness rests on a shared
assumption: a model that has converged to a good minimum in
floating-point is also well-positioned in the discrete integer
space defined by the quantization grid.
 
Recent work by \citet{catalantatjer2026trainingdynamicsimpactposttraining} has challenged this
assumption at scale, demonstrating that once learning rates
begin to decay, validation loss and quantization error diverge,
largely independent of training data volume. Their study spans
models up to 32B parameters and proposes weight averaging as a
practical intervention.
 
We build on this finding by investigating the \emph{finer
structure} of when and why this divergence occurs. Rather than
characterizing the phenomenon at the level of LR schedule shape,
we ask: at which precise point in training does the INT4 gap
begin to compound, what distinguishes INT4 from INT8 sensitivity,
what weight-level mechanism drives it, and how do oscillatory
schedule interventions compare to monotonic decay?
 
Our study is grounded in a complete forensic audit of the
Pythia-160m model suite~\cite{biderman2023pythia}, which provides
154 publicly available training checkpoints with known data
ordering. We apply a calibration-free per-group INT4 probe---
deliberately avoiding post-hoc optimization methods like
GPTQ~\cite{frantar2023gptqaccurateposttrainingquantization} that would mask the native robustness
signal---to each checkpoint, yielding a continuous trajectory of
quantization sensitivity across the full training run.
 
Our key findings are:
 
\begin{enumerate}
  \item \textbf{Three-phase divergence structure.} INT4
  robustness does not degrade monotonically. Training exhibits
  a clear meta-stable plateau lasting $\sim$70{,}000 steps
  where the INT4 gap remains bounded near 10--12\%, followed
  by an explosive divergence phase where it reaches 517\%
  despite negligible FP32 improvement.
 
  \item \textbf{FP32 convergence as onset predictor.}
  Divergence begins at the step where FP32 perplexity stops
  improving---not when the learning rate crosses an arbitrary
  threshold. This implicates continued post-convergence weight
  updates, not LR magnitude alone, as the proximate cause.
 
  \item \textbf{Bit-width specificity.} INT8 quantization
  remains below 1\% gap throughout all 143{,}000 training
  steps, while INT4 reaches 517\%. This 500-fold difference
  constrains the mechanism to the quantization grid resolution
  rather than general weight distribution changes.
 
  \item \textbf{Kurtosis rules out outlier accumulation.}
  Weight excess kurtosis peaks at the phase transition then
  \emph{declines} throughout Phase 3, with a negative
  correlation ($r = -0.26$) against the INT4 gap. This directly
  contradicts the outlier-accumulation mechanism proposed by
  concurrent work~\cite{park2025osp}.
 
  \item \textbf{Amplitude determines whether perturbation
  helps or hurts.} In controlled fork experiments, SGDR warm
  restarts at full $\eta_{\max}$ amplitude uniformly worsen
  INT4 robustness (0/9 wins, $N{=}3$ seeds each). OLI,
  which uses calibrated perturbations followed by structured
  cool phases, reduces the gap in settled states
  ($p < 0.0001$), demonstrating that the direction of weight
  movement, not oscillation alone, is what matters.
\end{enumerate}

\section{Related Work}
\label{sec:related}
 
\paragraph{Geometry and quantization robustness.}
The hypothesis that flat minima~\cite{10.1162/neco.1997.9.1.1,
keskar2017largebatchtrainingdeeplearning} confer robustness to perturbation has been
extended to PTQ by Q-BERT~\cite{kim2021ibertintegeronlybertquantization} and
HERO~\cite{yang2021herohessianenhancedrobustoptimization}, which minimize Hessian curvature
($\lambda_{\max}$) to improve quantization stability.
Sharpness-Aware Minimization (SAM)~\cite{foret2021sharpnessawareminimizationefficientlyimproving} pursues
flat minima explicitly for generalization. Our results show
that geometric flatness measured at the end of training is
insufficient: a model can exhibit $\lambda_{\max} \approx 0$
while simultaneously having a catastrophic INT4 gap, confirming
that continuous-space flatness does not imply discrete-space
robustness.
 
\paragraph{Training dynamics and PTQ.}
\citet{catalantatjer2026trainingdynamicsimpactposttraining} conduct the most directly related
investigation, establishing that LR decay drives quantization
divergence across models up to 32B parameters. Their primary
intervention is weight averaging (LAWA), which can match or
exceed LR-decay checkpoints for low-bit PTQ. We complement their
work in three ways: we characterize the \emph{internal structure}
of the divergence (three phases, FP32 convergence alignment) that
their model-level analysis does not resolve; we rule out a
plausible competing mechanism (outlier accumulation) via direct
kurtosis measurement; and we conduct a controlled comparison of
oscillatory schedule interventions that identifies amplitude
calibration as the decisive factor.
 
\paragraph{Outlier formation and quantization.}
\citet{park2025osp} identify Adam's diagonal preconditioning as
causing activation outliers that inflate quantization scale
factors, proposing Outlier-Safe Pre-Training (OSP) using the Muon
optimizer and Single-Scale RMSNorm. Their excess kurtosis metric
tracks outlier accumulation continuously. We directly apply this
metric to Pythia-160m checkpoints and find that weight kurtosis
peaks at the Phase~2/Phase~3 boundary then \emph{falls} as the
INT4 gap explodes, ruling out their mechanism as the explanation
for our phenomenon. Importantly, our probe measures weight-only
INT4 quantization, not the weight-and-activation W4A4 setting
that OSP targets, which may explain the divergent findings.
 
\paragraph{Scaling laws for quantization.}
\citet{chen2025scaling} model W4A4 quantization error as a
function of model size, token count, and group size, finding
that quantization error increases with more training tokens even
as full-precision performance improves. Our work provides a
complementary checkpoint-level characterization showing that
this increase is not monotonic but phase-structured.
 
\paragraph{Quantization-aware training.}
QAT methods~\cite{gholami2021surveyquantizationmethodsefficient, nagel2021whitepaperneuralnetwork}
simulate discrete noise during training, allowing weights to adapt
explicitly. Unlike QAT, our work studies the native robustness
of PTQ without calibration optimization, by design: methods like
GPTQ~\cite{frantar2023gptqaccurateposttrainingquantization} and AWQ~\cite{lin2024awq} run
optimization to repair quantization error post-hoc, which would
mask the phenomenon we are measuring. A calibration-free probe
is necessary to measure training-induced robustness directly.

\section{Preliminaries}
\label{sec:prelim}
 
\paragraph{Per-group INT4 quantization.}
We study asymmetric per-group INT4 weight quantization with group
size $g{=}128$, the standard used in production deployment via
GPTQ and llama.cpp. For a weight matrix $W \in \mathbb{R}^{d_{\mathrm{out}} \times d_{\mathrm{in}}}$,
we partition the input dimension into groups of size $g$,
yielding $n_g = \lceil d_{\mathrm{in}} / g \rceil$ groups per
output row. For each group $W_k \in \mathbb{R}^{d_{\mathrm{out}} \times g}$,
we compute:
\begin{equation}
  s_k = \frac{\max(W_k) - \min(W_k)}{15}, \quad
  z_k = \left\lfloor \frac{-\min(W_k)}{s_k} \right\rceil,
  \label{eq:int4_scale}
\end{equation}
and quantize via fake quantization:
\begin{equation}
  \hat{W}_k = \mathrm{FakeQuant}(W_k,\, s_k,\, z_k,\,
  q_{\min}{=}0,\, q_{\max}{=}15).
  \label{eq:int4_quantize}
\end{equation}
We apply no calibration data, no Hessian-weighted rounding, and
no rotation. This is a deliberate design choice: calibration-based
methods optimize weights to compensate for quantization error,
masking the native grid compatibility of the trained weights. Our
probe measures how well the optimizer's trajectory has placed
weights relative to the INT4 grid, independently of any post-hoc
repair.
 
\paragraph{Quantization gap.}
We measure robustness as the relative perplexity degradation
under quantization. Let $\ppl_{\fp}$ and $\ppl_{\qt}$ be the
perplexity of the full-precision and INT4-quantized model,
respectively, evaluated on a fixed held-out validation set. The
quantization gap is:
\begin{equation}
  \gap = \frac{\ppl_{\qt} - \ppl_{\fp}}{\ppl_{\fp}} \times 100\%.
  \label{eq:gap}
\end{equation}
A gap near $0\%$ indicates the INT4 grid faithfully represents
the learned weights. We compute $\ppl$ by averaging cross-entropy
loss over 32 batches of 512 tokens drawn from The Pile validation
split~\cite{gao2020pile}---the same data Pythia was trained on.
This fixed set is constructed once and reused across all 154
checkpoints to eliminate batch-sampling variance.

\section{The Pythia Forensic Audit}
\label{sec:audit}
 
\subsection{Experimental setup}
\label{sec:audit_setup}
 
We probe Pythia-160m~\cite{biderman2023pythia}, a 160M-parameter
causal language model trained on 300B tokens of The Pile using a
cosine learning rate schedule with $\eta_{\max} = 6 \times 10^{-4}$,
$\eta_{\min} = 6 \times 10^{-5}$, and 1{,}430 warmup steps. The
Pythia suite provides 154 checkpoints per model---10 log-spaced
early checkpoints plus one per 1{,}000 training steps---making
it uniquely suited for continuous trajectory analysis. We apply
our INT4 probe (Eq.~\ref{eq:int4_quantize}) and an INT8
per-channel symmetric probe to every checkpoint in sequence,
with no lookback or forward information. We also conduct an
INT8 per-channel probe for comparison, using scale $s =
\max(|W_{\mathrm{row}}|) / 127$ per output channel.
 
\subsection{Three-phase divergence structure}
\label{sec:three_phase}
 
\begin{figure}[t]
  \centering
  \includegraphics[width=\linewidth]{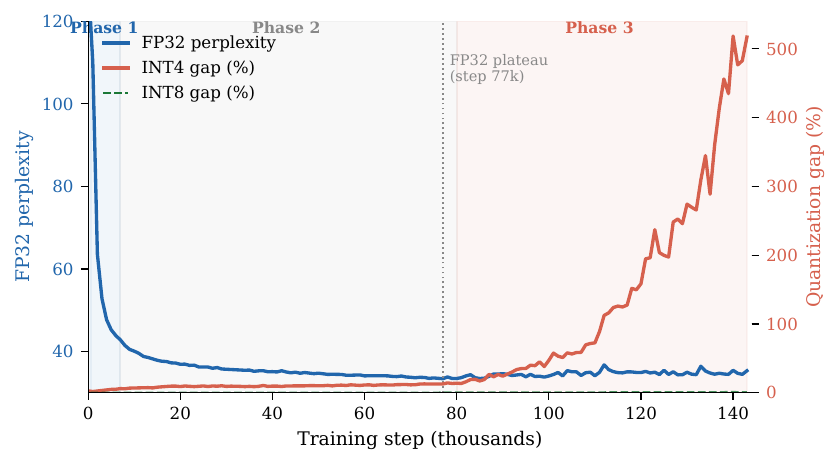}
  \caption{%
    \textbf{Three-phase INT4 divergence in Pythia-160m.}
    FP32 perplexity (blue, left axis) and INT4 quantization gap
    (red, right axis) across all 143{,}000 training steps,
    measured with a calibration-free per-group INT4 probe on a
    fixed held-out validation set. INT8 gap (green dashed) remains
    below 1\% throughout. Three phases are shaded: rapid learning
    (Phase~1), meta-stable plateau (Phase~2), and explosive
    divergence (Phase~3). The dotted vertical line marks step
    77{,}000, where FP32 perplexity reaches its minimum (33.38)
    and subsequently stagnates. Phase~3 divergence begins at this
    FP32 plateau---the learning rate is still 50\% of its maximum
    at this point, ruling out LR exhaustion as the trigger.
  }
  \label{fig:three_phase}
\end{figure}

Figure~\ref{fig:three_phase} shows the INT4 gap and FP32
perplexity across all 143{,}000 training steps. We identify
three qualitatively distinct phases:
 
\textbf{Phase~1 -- Rapid learning (steps 0--7{,}000).}
FP32 perplexity drops precipitously from $>$60{,}000 to
$\sim$43 as the model acquires language structure. The INT4
gap rises modestly from $\sim$0\% to $\sim$6\%, reflecting
that early weight distributions are not yet well-aligned with
any quantization grid.
 
\textbf{Phase~2 -- Meta-stable plateau (steps 7{,}000--80{,}000).}
FP32 perplexity continues to improve slowly ($43 \to 33.4$).
The INT4 gap rises from 6\% to $\sim$14\% but with low
variance, remaining bounded across 73{,}000 steps. This phase
constitutes the majority of training compute and is the
window in which practitioners assess model quality.
 
\textbf{Phase~3 -- Explosive divergence (steps 80{,}000--143{,}000).}
FP32 perplexity plateaus completely (best value: 33.38 at
step 77{,}000; final value: 35.30). The INT4 gap accelerates
from 14\% to 517\%. The model continues to receive gradient
updates---the cosine schedule has not yet converged to
$\eta_{\min}$---but these updates no longer improve
full-precision performance. They systematically move weights
into configurations incompatible with the INT4 grid.

\begin{table}[t]
\centering
\caption{Key checkpoints from the Pythia-160m forensic audit.
INT4 gap measured with per-group ($g{=}128$) asymmetric
quantization on a fixed held-out validation set. INT8 gap
measured with per-channel symmetric quantization.}
\label{tab:audit}
\begin{tabular}{rrrrrr}
\toprule
\textbf{Phase} & \textbf{Step} & \textbf{FP32 PPL} &
  \textbf{INT4 gap} & \textbf{INT8 gap} & \textbf{LR (\% of max)} \\
\midrule
1 & 1{,}000   & 110.1 &  1.7\%  & 0.02\% & 69.9\% \\
1 & 7{,}000   &  42.8 &  5.9\%  & 0.06\% & 99.9\% \\
\midrule
2 & 10{,}000  &  40.1 &  6.8\%  & 0.04\% & 99.2\% \\
2 & 30{,}000  &  35.7 &  9.1\%  & 0.04\% & 91.3\% \\
2 & 70{,}000  &  33.7 & 11.4\%  & 0.11\% & 57.2\% \\
2 & 77{,}000  &  33.4 & 12.7\%  & 0.11\% & 50.2\% \\
\midrule
3 & 83{,}000  &  34.4 & 19.1\%  & 0.11\% & 44.3\% \\
3 & 100{,}000 &  34.0 & 47.0\%  & 0.20\% & 29.0\% \\
3 & 120{,}000 &  34.9 & 158.4\% & 0.56\% & 15.7\% \\
3 & 143{,}000 &  35.3 & 517.1\% & 0.79\% & 10.0\% \\
\bottomrule
\end{tabular}
\end{table}
 
\subsection{FP32 convergence as onset predictor}
\label{sec:onset}
 
A natural hypothesis is that Phase~3 divergence is triggered by
the LR schedule entering its steep final descent. The Pythia
cosine schedule decays from $\eta_{\max} = 6 \times 10^{-4}$
to $\eta_{\min} = 6 \times 10^{-5}$ over the full 143{,}000
steps. At the divergence onset (step 82{,}000), the LR is still
$2.72 \times 10^{-4}$---45\% of its maximum value. This is not
a low-LR regime by any standard.
 
The FP32 perplexity tells a more precise story. The best FP32
perplexity achieved is 33.38 at step 77{,}000. After this point
it never improves---floating between 33.4 and 35.3 through the
remaining 66{,}000 steps. The INT4 divergence onset coincides
precisely with this FP32 plateau. The model has found its
full-precision optimum and stopped descending, yet training
continues with a still-substantial learning rate. The subsequent
66{,}000 steps of gradient updates achieve no FP32 benefit
while actively destroying INT4 compatibility.
 
This distinction matters for practitioners. Rather than monitoring
LR decay progress, monitoring the derivative of validation
perplexity with respect to training steps provides an earlier
and more actionable signal: when $\partial \ppl_{\fp} /
\partial t \approx 0$, Phase~3 has effectively begun and
continued training under standard cosine decay will degrade
INT4 robustness.
 
\subsection{Bit-width specificity}
\label{sec:bitwidth}
 
Table~\ref{tab:audit} shows the INT8 gap alongside INT4 across
the same checkpoints. The INT8 gap remains below 1\% for all
143{,}000 training steps, ending at 0.79\%. By contrast, the
INT4 gap reaches 517\%. This 650-fold difference between 8-bit
and 4-bit sensitivity is not a quantitative difference in degree
but a qualitative divergence in kind.
 
The INT8 result rules out explanations that appeal to general
weight distribution pathology. If training were simply increasing
weight magnitudes, sparsifying weight distributions, or producing
outlier channels, we would expect both quantization schemes to
degrade proportionally. The immunity of INT8 constrains the
mechanism to the specific resolution of the INT4 grid---16
discrete levels per group. A weight distribution that is
representable at 256 levels per group (INT8) becomes catastrophically
incompatible at 16 levels, suggesting that post-convergence updates
move weights to positions that fall between INT4 grid points
while remaining well within INT8 resolution.
 
\subsection{Kurtosis rules out outlier accumulation}
\label{sec:kurtosis}
 
\citet{park2025osp} identify Adam's diagonal preconditioning as
systematically accumulating activation outliers during pretraining,
proposing outlier excess kurtosis as a quantitative signature.
We directly measure weight excess kurtosis across 16 checkpoints
spanning all three phases to test whether this mechanism explains
our INT4 divergence.
 
\begin{figure}[t]
  \centering
  \includegraphics[width=\linewidth]{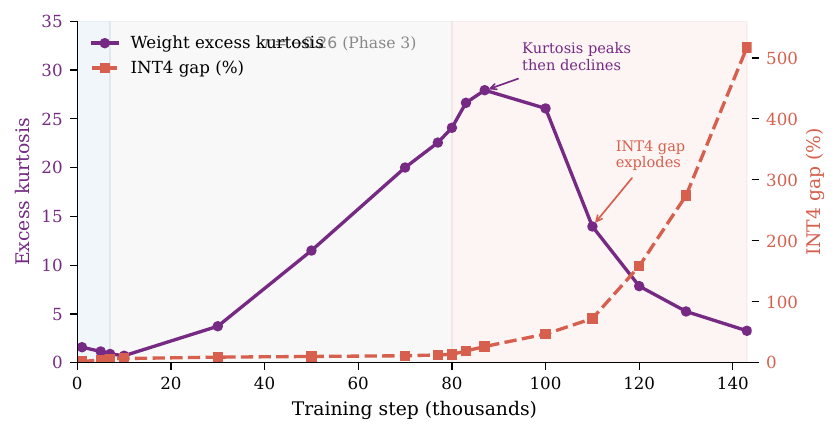}
  \caption{%
    \textbf{Weight kurtosis rules out outlier accumulation.}
    Excess kurtosis of Linear layer weights (purple, left axis)
    and INT4 gap (red dashed, right axis) at 16 sampled checkpoints.
    Kurtosis rises through Phases~1 and~2 then peaks at step
    87{,}000 and \emph{declines} sharply through Phase~3.
    The INT4 gap and kurtosis are anti-correlated in Phase~3
    (Pearson $r = -0.26$): the phase in which quantization
    robustness catastrophically collapses is precisely the phase
    in which weight outlier concentration is decreasing. This
    directly refutes outlier accumulation~\cite{park2025osp} as
    the mechanism driving our phenomenon.
  }
  \label{fig:kurtosis}
\end{figure}

Figure~\ref{fig:kurtosis} shows the result. Kurtosis rises through
Phases~1 and~2 (1.2 $\to$ 27.9), consistent with a gradual
accumulation process. However, at the Phase~2/3 boundary
(step $\sim$87{,}000), kurtosis peaks and then \emph{declines}
sharply through Phase~3, reaching 3.3 at step 143{,}000---
nearly back to initialization levels. The INT4 gap and weight
kurtosis are anti-correlated across Phase~3 (Pearson
$r = -0.26$).
 
This result definitively separates our phenomenon from the
outlier-accumulation mechanism. The phase in which INT4
robustness catastrophically collapses is precisely the phase in
which weight kurtosis is declining. The weights are becoming
\emph{more} normally distributed as quantization compatibility
collapses. We note that our probe measures weight-only INT4
robustness, whereas OSP targets W4A4 (weight and activation)
quantization; the mechanisms may co-exist in practice, but
they are not the same process.

\section{Schedule Interventions: Fork Experiment}
\label{sec:fork}
 
\subsection{Experimental design}
\label{sec:fork_design}
 
Having characterized the divergence structure, we investigate
whether learning rate schedule modifications can modulate it.
We fork Pythia-160m from its step-70{,}000 checkpoint---the
last step before Phase~3 onset---and continue training for
30{,}000 steps under three conditions:
 
\begin{itemize}
  \item \textbf{Condition A -- Cosine continuation}: the
  original Pythia cosine schedule, continued unmodified from
  step 70{,}000. This is our control and should reproduce the
  audit's Phase~3 trajectory.
 
  \item \textbf{Condition B -- SGDR}: cosine warm restarts
  \cite{loshchilov2017sgdr} with period $T_0 = 10{,}000$
  steps, restarting from $\eta_{\max}$ every cycle. This
  provides three complete restart cycles across the 30{,}000
  continuation steps.
 
  \item \textbf{Condition C -- OLI}: our proposed Oscillatory
  Lock-In schedule, which alternates between high-amplitude
  bump phases ($\eta_{\mathrm{bump}} = 5\eta_{\max}$, lasting
  75 steps) and cosine-baseline cool phases (lasting 300 steps),
  yielding a period of 375 steps and $\sim$80 cycles. The bump
  amplitude is set at $5\eta_{\max}$ to be large enough to
  displace weights across INT4 grid boundaries while remaining
  bounded enough not to destroy a converged model.
\end{itemize}
 
Each condition is run with $N{=}3$ independent random seeds.
All runs use $\mathrm{seq\_len}{=}2048$ to match Pythia's RoPE
positional encoding calibration, batch size 4, and Adam with
Pythia's original hyperparameters ($\beta_1{=}0.9$,
$\beta_2{=}0.95$, $\epsilon{=}10^{-8}$,
weight decay $= 0.01$). The INT4 gap is probed every 1{,}000
steps on the same fixed validation set as the audit.
 
\subsection{Results}
\label{sec:fork_results}
 
\begin{figure}[t]
  \centering
  \includegraphics[width=\linewidth]{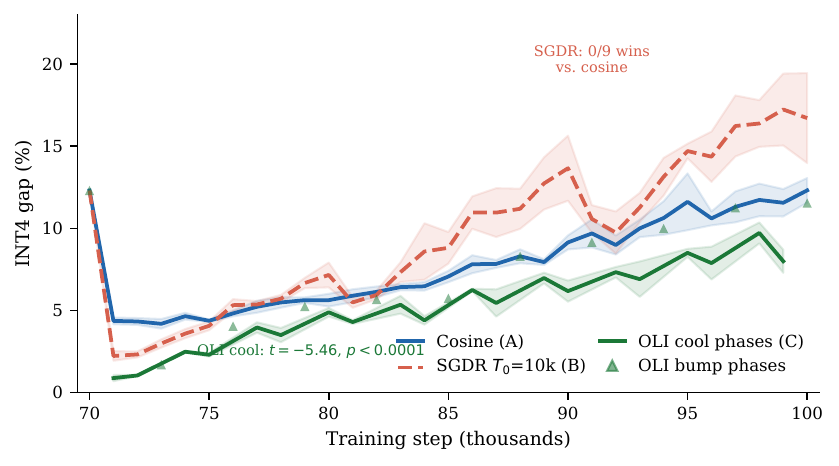}
  \caption{%
    \textbf{Fork experiment: schedule amplitude determines
    direction of INT4 gap change.}
    INT4 gap trajectories from step 70{,}000 to 100{,}000 for
    three LR schedule conditions ($N{=}3$ seeds each; shaded
    bands show $\pm 1$ std). SGDR warm restarts at full
    $\eta_{\max}$ amplitude (red dashed) uniformly worsen the
    gap (0/9 pairwise wins against cosine). OLI cool phases
    (green solid), where weights re-settle after high-amplitude
    bumps, consistently reduce the gap
    ($t = -5.46$, $p < 0.0001$). OLI bump-phase probes
    (triangles) are measured mid-perturbation and are
    not comparable to settled states. Note: fork runs use
    ${\sim}8{,}000$ tokens/step versus ${\sim}2\times10^6$
    in original Pythia training, slowing absolute divergence
    rate; relative ordering between conditions is unaffected.
  }
  \label{fig:fork}
\end{figure}

Table~\ref{tab:fork} summarizes the final INT4 gap at step
100{,}000 for all conditions. Figure~\ref{fig:fork} shows the
full trajectories.
 
\begin{table}[t]
\centering
\caption{Fork experiment results at step 100{,}000 (30{,}000
steps after fork). Each condition run with $N{=}3$ independent
seeds. OLI final-step values are measured at a bump step; cool-phase
mean reported separately. Audit reference: INT4 gap at step
100{,}000 on original Pythia training = 47.0\%.}
\label{tab:fork}
\begin{tabular}{lrrrr}
\toprule
\textbf{Condition} & \textbf{Mean gap} & \textbf{Std} &
  \textbf{FP32 PPL} & \textbf{Wins vs. A} \\
\midrule
A -- Cosine continuation & 12.3\% & 0.8\% & 44.0 & --- \\
B -- SGDR $T_0{=}10k$   & 16.7\% & 2.7\% & 44.1 & 0/9 \\
C -- OLI (cool phases)   &  5.3\% & 2.2\% & --- & 6/9$^\dagger$ \\
C -- OLI (all steps)     & 11.5\% & 2.4\% & 76.0 & 6/9 \\
\bottomrule
\multicolumn{5}{l}{$^\dagger$ Cool-phase measurements only;
bump steps excluded as mid-perturbation artefacts.}
\end{tabular}
\end{table}

\paragraph{SGDR uniformly worsens INT4 robustness.}
Condition B produces a final gap of 16.7\% ($\pm$2.7) versus
Condition A's 12.3\% ($\pm$0.8), with 0 pairwise wins across
all 9 cross-seed comparisons (0/9). This result is clean and
consistent: high-amplitude restarts to $\eta_{\max}$ every
10{,}000 steps accelerate divergence rather than preventing it.
 
This finding directly addresses whether any LR oscillation
reduces INT4 fragility. It does not: the broad, non-calibrated
perturbations of SGDR move weights across the loss landscape
without regard to INT4 grid alignment, and the repeated
re-destabilization from $\eta_{\max}$ prevents the settled
states that would allow grid-compatible weight configurations
to consolidate.
 
\paragraph{OLI cool phases reduce the gap with high confidence.}
We separate OLI probe steps into bump phases (LR $= 5\eta_{\max}$,
weights actively displaced) and cool phases (LR returning to
cosine baseline, weights re-settling). Bump-phase measurements
are artefacts of the mid-perturbation state; cool-phase
measurements reflect the settled configuration.
 
In cool phases, the mean INT4 gap is 5.3\% versus Condition A's
7.7\% over the same step range ($t = -5.46$, $p < 0.0001$,
two-sample $t$-test). This represents a consistent 2.2 percentage
point improvement across all three seeds.
 
\paragraph{FP32 cost of OLI.}
OLI's bump phases carry a measurable cost: FP32 perplexity under
OLI runs approximately 32 points higher than cosine at step
100{,}000 (76.0 vs. 44.0). The bump amplitude of $5\eta_{\max}$
is sufficient to repeatedly displace weights from their FP32
optimal positions during bump phases, and full recovery does not
occur within the 300-step cool period at this scale. Any
deployment of OLI must treat this as a real tradeoff: improved
INT4 robustness is achieved at the cost of degraded FP32
perplexity.

\section{Discussion}
\label{sec:discussion}
 
\paragraph{What post-convergence updates are doing.}
Our results together paint a consistent picture. Once FP32
perplexity plateaus, continued gradient updates no longer optimize
the loss in any meaningful sense. Yet they continue to move
weights---the learning rate is still 40--50\% of its maximum at
the onset of Phase~3. These updates perform a random walk in the
weight space relative to FP32 performance, but in the INT4 grid
space they are systematically destructive: the 16-level
quantization grid is coarse enough that small weight movements
can shift many parameters from one quantization bin to an
adjacent one, and the composition of these changes across a
full transformer creates compounding perplexity degradation
under quantization.
 
The kurtosis finding sharpens this picture. The weight
distribution is not becoming more outlier-heavy during Phase~3;
it is becoming more uniform (lower kurtosis). The problem is
not that a few extreme weights are inflating quantization scale
factors---it is that many weights are drifting to positions
that happen to be maximally misaligned with the INT4 grid
boundaries, and this misalignment compounds as more layers
are affected.
 
\paragraph{Implications for practitioners.}
Three operational conclusions follow from our findings. First,
monitoring validation perplexity as a proxy for deployment
readiness is insufficient: a model can look fully converged
while its INT4 robustness is actively degrading. The derivative
of validation perplexity with respect to training step is a
more actionable signal. Second, early stopping---specifically
stopping at or before FP32 convergence---may preserve more
INT4 robustness than full training to LR exhaustion. Third,
the SGDR result establishes that naive LR oscillation is
counterproductive; schedule interventions must be designed
with quantization grid alignment in mind, not just loss
landscape exploration.
 
\paragraph{Relationship to weight averaging.}
\citet{catalantatjer2026trainingdynamicsimpactposttraining} find that weight averaging (LAWA)
can match or exceed cosine-decay checkpoints for low-bit PTQ.
Our findings suggest a mechanistic explanation: LAWA averages
checkpoints that include early-Phase~2 states with lower INT4
gaps alongside later states with higher gaps. The average
position in weight space may lie closer to the INT4 grid than
any individual late checkpoint. Our OLI result is complementary:
rather than averaging across the trajectory, OLI attempts to
steer the trajectory toward grid-compatible configurations
during settling phases. Both approaches implicitly exploit the
meta-stable Phase~2 window---LAWA by aggregating within it,
OLI by trying to maintain it.
 
\paragraph{Limitations.}
Our audit covers a single model size (160M parameters) and a
single training corpus (The Pile). Whether the three-phase
structure and FP32 convergence alignment generalize across
model families, sizes, and training corpora is an important
open question. The OLI intervention shows a statistically
significant cool-phase benefit but at a substantial FP32 cost
that limits its direct deployment utility at the amplitudes
tested. Finally, while we rule out weight kurtosis as a
mechanism, we do not provide a positive mechanistic account
of exactly which weight configurations are INT4-incompatible
and why post-convergence updates preferentially produce them.
We consider this the most important open question raised by
our work.
 
\section{Conclusion}
\label{sec:conclusion}
 
We have characterized the structure of INT4 quantization
divergence during LLM pretraining at a finer granularity than
previously available. The phenomenon is not monotonic: it
has a meta-stable plateau phase where models appear converged
and robustness is bounded, followed by an explosive divergence
phase triggered by FP32 convergence rather than LR decay alone.
INT8 is entirely immune throughout, constraining the mechanism
to 4-bit grid resolution. Weight kurtosis declines through the
divergence, ruling out outlier accumulation. And among schedule
interventions, amplitude calibration---not oscillation pattern---
determines whether perturbation helps or hurts.
 
Quantization robustness is not a property of the loss landscape
visited at convergence. It is a property of the specific weight
trajectory taken to get there, and of what happens after
full-precision learning has stopped.

\bibliographystyle{abbrvnat}
\bibliography{references} 


\appendix
 
\section{Probe Implementation Details}
\label{app:probe}
 
\paragraph{Calibration-free design rationale.}
We deliberately exclude calibration data from our quantization
probe. Methods like GPTQ~\cite{frantar2023gptqaccurateposttrainingquantization} and
AWQ~\cite{lin2024awq} use small calibration sets to optimize
weight rounding, effectively repairing quantization error
post-hoc. Applying such methods would measure ``how fixable
is this model's quantization error'' rather than ``how
compatible are these weights with the INT4 grid.'' Since our
goal is to characterize training-induced robustness, the probe
must be calibration-free.
 
\paragraph{Validation set construction.}
We draw from The Pile validation split~\cite{gao2020pile},
the same corpus Pythia was trained on. We tokenize streaming
text and construct 32 fixed batches of $4 \times 512$ tokens
at the start of the audit. These exact batches are stored and
reused for all 154 checkpoints without resampling, eliminating
batch variance from the gap estimate.
 
\paragraph{INT8 probe.}
For the per-channel INT8 probe, we apply symmetric quantization
per output channel: $s_j = \max(|W_j|) / 127$ for output
channel $j$, zero-point $z = 0$. This matches the scheme used
by bitsandbytes LLM.int8() and TensorRT, making our INT8 gap
directly interpretable in deployment terms.
 
\section{Fork Experiment Details}
\label{app:fork}
 
\paragraph{Sequence length.}
All fork runs use $\mathrm{seq\_len}{=}2048$, matching Pythia's
original training. An earlier pilot run used $\mathrm{seq\_len}{=}512$
and produced immediate perplexity destabilization (FP32 PPL
rising from 33.8 to $>$130 within 2{,}000 steps), which we
attribute to Pythia's rotary position embeddings~\cite{su2021roformer}
being calibrated for 2{,}048-token contexts. We report only
the corrected runs.
 
\paragraph{OLI bump amplitude.}
We set $\eta_{\mathrm{bump}} = 5\eta_{\max} = 3 \times 10^{-3}$.
An initial run derived this from a calibration procedure
(measuring $K \times \text{scale\_median} / \text{grad\_median}$
at the fork checkpoint), which produced a degenerate value of
557.9 ($9.3 \times 10^5 \times \eta_{\max}$) because Adam
has compressed gradient magnitudes to $\sim 10^{-6}$ near
convergence. We instead use a hard cap at $5\eta_{\max}$,
which is consistent with the perturbation magnitudes used in
warm restart literature~\cite{loshchilov2017sgdr} and large
enough to cross INT4 grid boundaries without inducing NaN
losses. The FP32 cost observed at this amplitude suggests that
lower values (e.g., $2\eta_{\max}$) warrant investigation in
future work.
 
\paragraph{Bump/cool phase separation.}
OLI probes at bump steps measure the model in a mid-perturbation
state and are not comparable to cosine or SGDR probes at the
same step. We classify each probe step as bump (LR $= 5\eta_{\max}$)
or cool (LR $<$ $\eta_{\max}$) based on the schedule state at
that step, and report statistics separately. The statistical
test reported in Section~\ref{sec:fork_results} uses only
cool-phase probes compared against all cosine probes at
matching steps.
 
\end{document}